\documentclass[11pt,a4paper]{article}
\usepackage[hyperref]{eacl2021}
\usepackage{times}
\usepackage{booktabs, multirow, array, graphicx} 
\usepackage[ruled,vlined, noend, linesnumbered]{algorithm2e}
\usepackage[export]{adjustbox}
\usepackage{latexsym}
\usepackage{amsmath, amsfonts}
\usepackage{enumerate}

\newcommand{\STAB}[1]{\begin{tabular}{@{}c@{}}#1\end{tabular}}
\newcommand{\halftick}{\checkmark\kern-1.1ex\raisebox{.7ex}{\rotatebox[origin=c]{125}{--}}}
\DeclareMathOperator*{\argmax}{arg\,max}

\usepackage{microtype}

\aclfinalcopy 
\def\aclpaperid{29} 

\title{Discrete Reasoning Templates for Natural Language Understanding}

\author{Hadeel Al-Negheimish \\
 
  \And
  Pranava Madhyastha \\\\
  Department of Computing \\
  Imperial College London \\
  \texttt{\{halnegheimish,pranava,a.russo\}@imperial.ac.uk} \\
  \And
  Alessandra Russo \\
  
 }

\date{}

\begin{document}
\maketitle
\begin{abstract}
 Reasoning about information from multiple parts of a passage to derive an answer is an open challenge for reading-comprehension models. In this paper, we present an approach that reasons about complex questions by decomposing them to simpler subquestions that can take advantage of single-span extraction reading-comprehension models, and derives the final answer according to instructions in a predefined reasoning template. We focus on subtraction based arithmetic questions and evaluate our approach on a subset of the DROP dataset. 
 We show that our approach is competitive with the state of the art while being interpretable and requires little supervision.
\end{abstract}

\section{Introduction}
\label{sec:intro}
\begin{figure}[ht]
\centering
\resizebox{0.9\columnwidth}{!}{
\includegraphics[width=0.89\columnwidth,keepaspectratio]{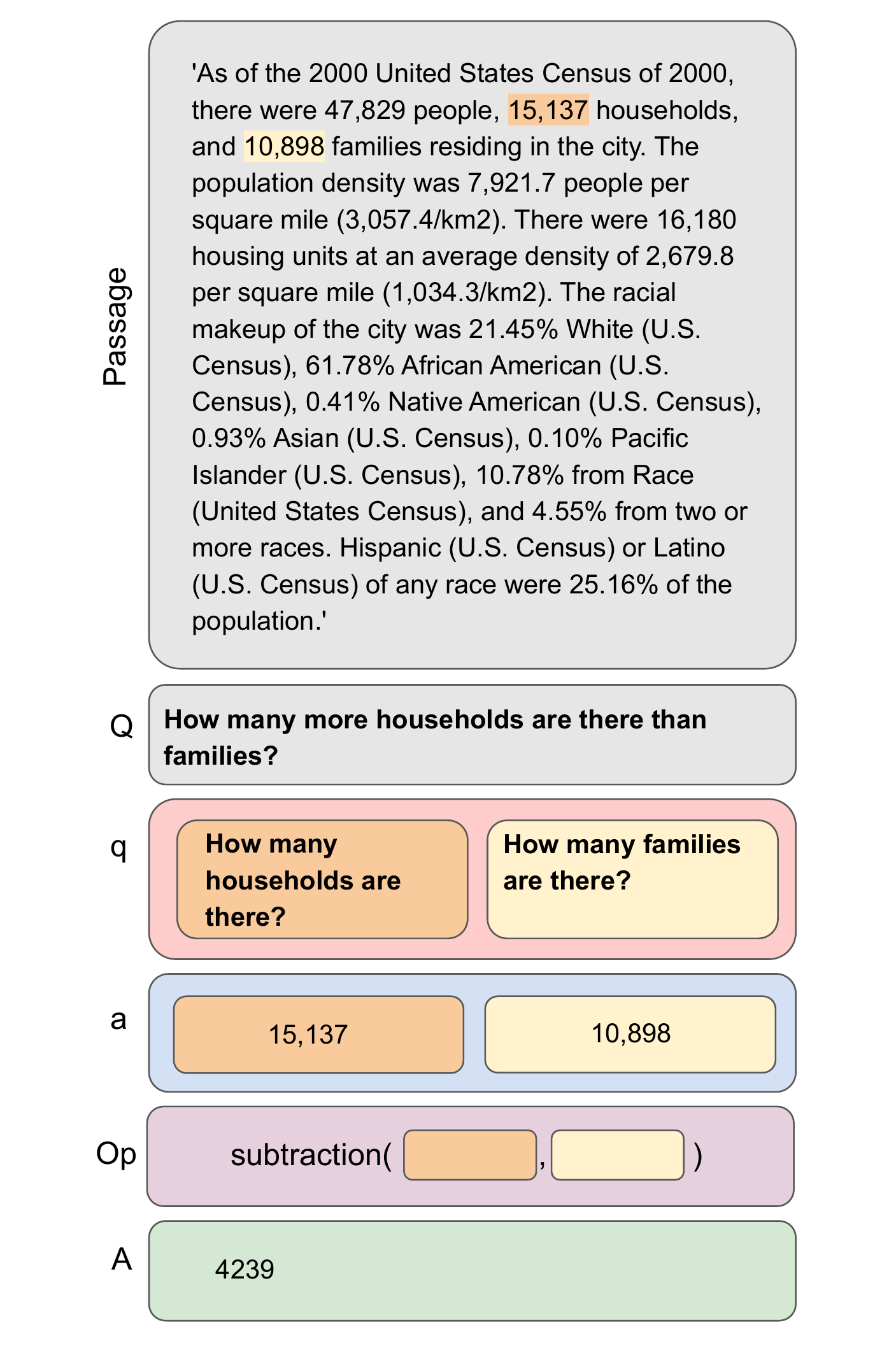}
}
\caption{Subtraction Template:
    Original question is decomposed to two simpler subquestions that find the values associated with the two compared entities as span-extraction, and the final answer is calculated by finding the absolute difference of the two partial answers.}
\label{fig:model-overviewl}
\end{figure}
Automated reading comprehension (RC) is an important natural language understanding task, where a model is presented with a passage and is asked to answer questions about that passage. While models have excelled at single-span extraction questions, they still struggle with reasoning over distinct parts of a passage \cite{Dua2019DROP}.
Several multi-hop reasoning benchmarks have been proposed \cite{yang2018hotpotqa,khashabi-etal-2018-looking, Dua2019DROP}, of which, in this paper, we focus on the DROP (Discrete Reasoning Over the content of Paragraphs) dataset. Inspired by the semantic parsing literature,  the dataset contains questions that involve possibly multiple steps of discrete reasoning over the contents of paragraphs, including numerical reasoning. 

Recent work has proposed several novel approaches to tackle DROP \cite{ran-etal-2019-numnet, hu-etal-2019-multi, andor-etal-2019-giving, Gupta2020Neural, Chen2020Neural}. However, most approaches provide little evidence of their reasoning process, especially with regards to \emph{why} specific operands are chosen for a reasoning task. With the exception of \cite{Gupta2020Neural, Chen2020Neural}, they also suffer from limited compositionality. 

In this paper, we present a first attempt at building an interface between discrete reasoning and unstructured natural language. We propose decomposing a question to simpler subquestions that can more easily be solved by single-span extraction RC models. Such decomposition is defined by Reasoning Templates, which also determine how to assemble the computed partial answers. We demonstrate the feasibility of our approach with the \emph{subtraction} based questions (illustrated in Figure~\ref{fig:model-overviewl}). We show that our approach is competitive with the state of the art models on a subset of DROP’s subtraction questions while requiring much less training data and providing visibility of the model’s decision-making process.


\section{Related Work}
\label{sec:rel-work}
There has been a recent resurgence in research on automated reading comprehension (RC) where an automated system is capable of reading a document in order to answer the questions pertaining to the document. This has led to the creation of several RC datasets to facilitate the research \cite{rajpurkar-etal-2016-squad, rajpurkar2018know, yang2018hotpotqa, reddy-etal-2019-coqa, Dua2019DROP, huang-etal-2019-cosmos}.
Among these, SQuAD \cite{rajpurkar-etal-2016-squad} is a popular \emph{single-hop question answering} dataset where a question can be answered by relying on a single sentence from the document. 
SoTA models have achieved near-human performance on such single-hop question answering tasks.\footnote{https://rajpurkar.github.io/SQuAD-explorer/} However, answering a question by only identifying the most relevant span leaves models prone to exploiting advanced pattern matching algorithms.

On the other hand, multi-hop questions make reading-comprehension more challenging, as they require integrating information from multiple parts in a passage \cite{yang2018hotpotqa,khashabi-etal-2018-looking, Dua2019DROP}. DROP \cite{Dua2019DROP} is one such dataset that 
contains questions covering many types of reasoning, such as counting, sorting, or arithmetic. The dataset was constructed by adversarially crowdsourcing questions on a set of Wikipedia passages known to have many numbers. 
Special model architectures have been built to tackle DROP, and these fall into two general directions: the first direction augments reading comprehension models that were successful on single-span extraction questions with specialized modules that tackle more complex questions. These include NAQANet \cite{Dua2019DROP}, NumNet \cite{ran-etal-2019-numnet} , and MTMSN \cite{hu-etal-2019-multi}. The second direction works on predicting programs that would solve the question, CalBERT \cite{andor-etal-2019-giving} defines a set of derivations and scoring functions for each of them, while more recent work NMN \cite{Gupta2020Neural} and NeRd \cite{Chen2020Neural} utilize LSTMs to decode variable-length programs from question and passage embeddings. 

By definition, models with specialized modules have limited compositional reasoning abilities. 
The two directions vary in their interpretability; the first shows which module has been used, and the second shows the resulting programs which have been generated to compute the answer. However, none of these directions indicate why operands in the passage were selected. For all approaches, the dataset is augmented with all possible derivations that lead to the gold answer, by performing an exhaustive search. Moreover, all approaches assume a pre-processing step that extracts all numbers in the passage and their indices, which massively reduces the search space for arithmetic questions.

In this work, we build upon DecompRC \cite{min-etal-2019-multi} for question decomposition, where a model is trained to extract key parts of content from the question which are then used for decomposition.  Arithmetic questions, which we focus on in this work, are a known limitation of DecompRC.
An alternative approach to decomposition is QDMR \cite{Wolfson2020Break}, a recently proposed formalism for decomposing questions into a series of simpler steps based on predefined query operators.  QDMR breaks down a question to its atomic parts directly, whereas we propose recursively decomposing questions to simpler ones. While \cite{Wolfson2020Break} provides a dataset of annotated questions, QDMR parsing remains an open challenge.
In the following section we present our approach that focuses on answering arithmetic questions. 

\section{Approach}
\label{sec:approach}
We propose a pipelined approach that focuses on breaking down complex questions that require reasoning over multiple parts in the passage to simpler single-hop questions. The latter can be resolved by taking advantage of state of the art single-hop reading comprehension models.
The main building block of our approach is a \textit{reasoning template}. Each reasoning type is associated with a single template, which contains instructions on how to decompose a question and how to combine partial answers to arrive at the final answer. 

\begin{figure}[ht]
    \centering
\includegraphics[width=\columnwidth]{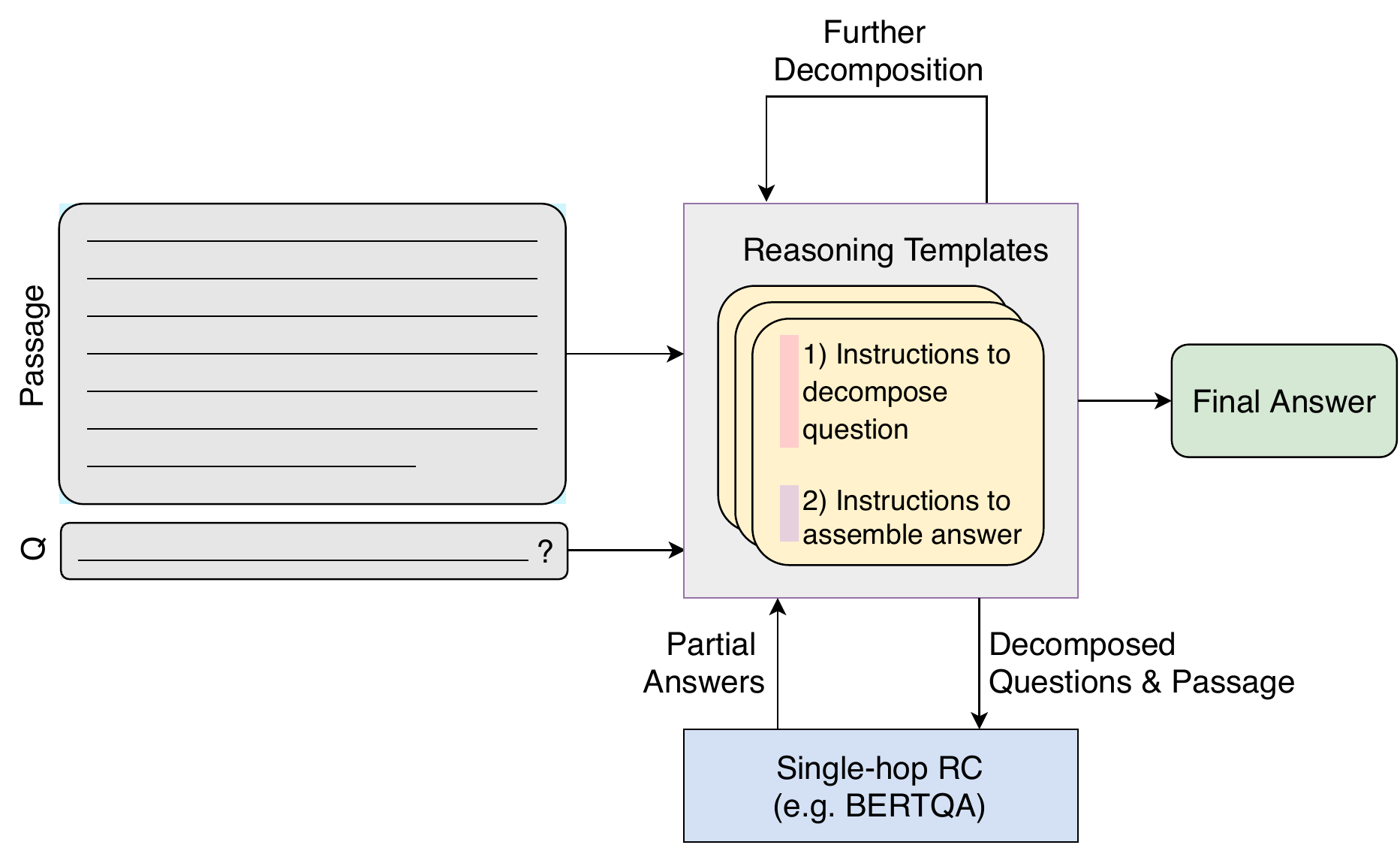}
    \caption{Model Overview: Given a question, decompose it into simpler questions according to a template such that they can be answered by a single-hop RC model, and assemble the final answer by applying the operation associated with the template. }
    \label{fig:reasoningtemplates}
\end{figure}
Figure~\ref{fig:reasoningtemplates} illustrates our pipeline. First, the question and passage are fed to our system, which selects a template  depending on the reasoning type required (classification task). The template decomposes the question to simpler subquestions that are then passed on to a single-hop RC model. Partial answers are used to arrive at an answer according to the instructions provided by the template. Some questions need further decomposition, and the appropriate template will be chosen for the subquestion.\footnote{Further decomposition and template selection modules are left for future research.}

For the question decomposition component, our approach closely follows and builds upon DecompRC  \cite{min-etal-2019-multi}, originally proposed for multihop, multidocument question answering. We repurpose the model for multihop arithmetic questions. 
DecompRC uses a two step approach to decompose questions. First, a pointer model is trained to identify a key part of the question that is used to formulate the sub-questions. Second, the predicted pointers are used to procedurally change the original question into two or three sub-questions. The approach is defined for three types of questions: Bridging, Intersection, and Comparison; each of them uses a different pointer model and a different heuristic procedure to generate the sub-questions.

In this paper, we propose the discrete reasoning template framework and demonstrate its potential by defining a single template:
\emph{subtraction}. We describe in detail our approach in the following subsections. 

\subsection{Question Decomposition}
\label{sec:QuestionDecomp}
Question decomposition is a two-step process that includes identifying relevant information in the text of the original question (span extraction), and then using those spans for heuristic generation of subquestions. The output of this step are simpler subquestions, see `$q$' in Figure~\ref{fig:model-overviewl}. 

\paragraph{Span Extraction}
 For subtraction questions, the spans we are interested in identify two entities whose associated values are to be subtracted.  Consider `$_0$How $_1$many $_2$more $_3$\textit{households} $_4$are $_5$there $_6$than \textit{$_7$married $_8$couples $_9$living $_{10}$together$_{11}$}?’. We need to extract the start and end indices of the first entity \textit{households}, and the start and end indices of the second entity \textit{married couples living together}, which are $[3,3,7,10]$.
 
The pointer model is trained to predict 4 pointers, the start and end indices of the first and second entity respectively. Concretely, the model extracts 4 indices, $p_1\leq p_2 \leq p_3 \leq p_4$, that surround the two spans of interest, maximizing the joint probability:
\[p_1,\cdots,p_4 = \argmax_{\{i_1\leq \cdots \leq i_4\}} \prod ^{4}_{j=1} \mathbb{P}(i_j=ind_j) \]

\noindent
where $\mathbb{P}(i_j=ind_j)=Y_{ij}$ is the probability that the $i$th word is the $j$th index produced by the pointer, and 
\[Y= \text{softmax}(UW) \in  \mathbb{R}^{n \times 4}\]
 where $W$ is a learned weight matrix of size $h \times 4$ and $U$ is the contextualized embeddings of length $h$ produced by pre-trained BERT\cite{devlin2018bert} of the $n$ tokens in the original question:
\[U= \text{BERT}(S) \in \mathbb{R}^{n \times h}\]
\noindent{We then train this model using cross entropy loss until convergence}. 

\paragraph{Subquestion Generation}
We find that the subquestions needed in our approach have a high degree of overlap with the original question, making them amenable to heuristic decomposition as in DecompRC \cite{min-etal-2019-multi}. While DecompRC is defined for bridging, intersection and comparison type questions, we extend it with a separate procedure to handle \textit{subtraction} type questions as described below.
We outline in Algorithm \ref{algo-subtraction} how subquestions can be generated for subtraction questions, given the pointers that have been predicted by the previous step. 
\begin{algorithm}[t]
\small
\SetAlgoLined
\KwData{Original question q: string, pointers $P_4$: array of length 4}
\KwResult{subquestions $q_1$,  $q_2$: strings}
 dep\_parse= dependency\_parse(q)\;
 part1 $\leftarrow$ q[0:$p_1$]\;
 ent1$\leftarrow$q[$p_1$:$p_2+1$]\;
 middle$\leftarrow$q[$p_2+1$:$p_3$]\;
 ent2$\leftarrow$q[$p_3$:$p_4+1$]\;
 part2$\leftarrow$ q[$p_4+1$:end]\;

 \For{word \textbf{in} part1}{
 \If{word.pos\_tag  \textbf{in} \texttt{[`JJR', `RBR']}}{remove word from part1\;}
 }

 head $\leftarrow$ dep\_parse.parent(ent2)\;
 i $\leftarrow$ head.i\;
 prev\_i $\leftarrow$ i\;
 \While{ (head in middle) \textbf{AND} (prev\_i-i $\leq$ 1)}{
 
new\_head $\leftarrow$ dep\_parse.parent(head) \;
  remove head from middle\;
  head $\leftarrow$ new\_head\;
  prev\_i $\leftarrow$ i\;
   i $\leftarrow$ head.i\;   
  
 }
 
 q1$\leftarrow$ part1+ent1+middle+part2\;
 q2$\leftarrow$ part1+ent2+middle+part2\;

 \caption{Subquestion generation for subtraction questions}
 \label{algo-subtraction}
\end{algorithm}
The algorithm keeps words that are common for both subquestions and then places each of the entities in the center of the generated questions. First, we chunk the original question into parts using the pointers as in lines 2-6. In lines 7-9, we remove comparative adjectives and adverbs from the first part. Before concatenating the different parts again, we remove the extra words from the middle part, utilizing the dependency parse of the original question.

\subsection{Single-hop question answering}
\label{sec:BERTQA}
Once we have decomposed questions into simpler, single-hop questions, we can use the subquestions to extract the appropriate operands for reasoning from the passage. We opt to make use of a pre-trained off-the-shelf single-span extraction model, details provided in section \ref{sec:exp}.
This is one possible instantiation for the model, and we can use any robust span-extraction model in its place.

\subsection{Operation}
\label{sec:operation}
A reasoning template includes instructions on how to perform two main steps; the first step decomposes a question to simpler subquestions as we have described in section~\ref{sec:QuestionDecomp}. The second step, operation, is designed to derive the final answer given partial answers to decomposed questions. In the case of subtraction, it is simply the absolute difference between the two retrieved values, see `\emph{Op}' in Figure~\ref{fig:model-overviewl}. In the case where a span retrieved for a decomposed question contains more than a single number, we use the first number in the span. 
\section{Experiments}
\label{sec:exp}
We start with a single template to demonstrate our approach: \textit{subtraction}. Subtraction questions rely on finding the difference between two numbers to find the answer, they are usually in the form of \textit{`How many more..?’} or \textit{`How many fewer..?’}.

\paragraph{Dataset}
For evaluation, we collect two sets of subtraction questions from the DROP development set. The first, \emph{clean}, is a subset of 52 questions curated  by filtering the original dataset to find questions that contain words with `JJR' or `RBR' pos-tags (comparative adjective and comparative adverb respectively), and from those we randomly sample 10 questions at a time and manually identify subtraction questions. 
We also annotate each of these questions with gold decompositions, two subquestions for each complex question. 
The other evaluation set, \emph{noisy},  is a larger dataset that has been heuristically generated, this is intended to support generalizability of results on the smaller evaluation set. It contains 892 questions that have been filtered using trigrams at the beginning of the question: \textit{`How many more’} or \textit{`How many fewer’}.

There are two learning components in our pipeline: a pointer model to extract relevant entities from the question and a single-hop RC model to answer decomposed questions. For the latter, we use an off-the-shelf pre-trained BERT \cite{devlin2018bert} question answering model, which has been fine-tuned on SQuAD \cite{rajpurkar-etal-2016-squad}, a single-hop reading comprehension dataset. Specifically, we use the one provided by the huggingface transformer library \cite{Wolf2019HuggingFacesTS}.
As for the former, to train the pointer model we follow \cite{min-etal-2019-multi} and annotate 200 examples. The data for this was gathered from the DROP training set in the same way we curated the \emph{clean} evaluation set, for this step we simply identify the compared entities and delimit them with `\#'. 

\subsection{Results and Discussion}
\label{sec:results}
\paragraph{Evaluating Question Decomposition}
In Table~\ref{table:pointer-accuracy} we report the accuracy of the pointer model on the \emph{clean} subtraction evaluation set, and in Table~\ref{table:f1-spans} we measure the overlap between the resulting spans and the annotated entities. While getting all pointers to match label succeeds for $73\%$ of the data, we note that the accuracy of each of the pointers is much higher. We find that the pointer delimiting the start of the first entity is seemingly the most difficult to predict, which is also seen in lower F1 score for the first entity. We conjecture this to be the likely case as the second entity is usually preceded by words such as `than' or `compared to'.
\begin{table}[ht]
\adjustbox{width=\columnwidth}{%
\begin{tabular}{@{}llllll@{}}
\toprule
         & p1    & p2    & p3    & p4    & all   \\ \midrule
Acc & $84.0 \pm 0.9$ & $88.5 \pm 1.6$ & $98.1$ & $94.9 \pm 0.9$ & $73.1$ \\ \bottomrule
\end{tabular}}
\caption{Accuracy of Pointer$_4$ model, we list the accuracy of individual pointers separately and accuracy of all pointers for each example. Results are reported as an average of 3 runs of the model with different random seeds.}
\label{table:pointer-accuracy}
\end{table}
\begin{table}[ht]
\normalsize
\centering
\begin{tabular}{@{}lll@{}}
\toprule
         & First Entity    & Second Entity      \\ \midrule
F1 & $0.89 \pm 0.02$ & $0.97 \pm 0.003$ \\
Precision  & $0.91 \pm 0.018$ & $0.96 $ \\
Recall  & $0.90 \pm 0.023$ & $0.99 \pm 0.006$ \\
\bottomrule
\end{tabular}
\caption{Measured overlap between resulting spans of the predicted pointers and the annotated entities, averaged over all questions in \emph{clean} evaluation set.}
\label{table:f1-spans}
\end{table}
\begin{table}[ht]
\centering
\begin{tabular}{@{}lll@{}}
\toprule
Similarity Measure     & q1     & q2     \\ \midrule
WMD$_{max}$            & $3.56$   & $4.43$   \\
WMD$_{avg}$            & $0.2266$ & $0.6714$ \\
WMD$_{median}$         & $0.0$    & $0.0$    \\
$cos(\theta)_{min}$    & $0.9538$ & $0.9476$ \\
$cos(\theta)_{avg}$    & $0.9959$ & $0.9913$ \\
$cos(\theta)_{median}$ & $1.0$    & $1.0$    \\ \bottomrule
\end{tabular}
\caption{Reported similarities between manually decomposed questions (gold) and decompositions generated by our approach. We use word mover's distance (WMD) and cosine similarity of average word embeddings. For the former we report \emph{max} distance, while in the latter we report \emph{min} similarity as these highlight the worst-case of all subquestions. For most examples, the gold decompositions and generated subquestions overlap perfectly, as indicated by \emph{median} score.  }
\label{table:distance-gold}
\end{table}
We also measure the similarity between decomposed questions generated by our approach and the manually annotated gold decompositions. 
Table \ref{table:distance-gold} displays the Word Mover's Distance metric \cite{Kusner-2015-WMD} and cosine similarity, based on the GloVe word embeddings shipped with SpaCy's \cite{spacy2} \texttt{en\_core\_web\_lg} model. For most questions, the two subquestions match perfectly between the gold annotations and the generated ones. However, upon manual inspection, we find that the generated subquestion might sometimes omit the final verb. This is because of our traversal of the dependency parse in Algorithm \ref{algo-subtraction}. We found BERTQA was robust to these differences when extracting the related span from the passage.

\paragraph{Evaluating the Approach}
\begin{table}[t]
\centering
\adjustbox{width=\columnwidth}{%
\begin{tabular}{@{}lllll|l@{}}
\toprule
& Model & Acc$_{c}$ & Acc$_{c}^{-}$ &  \# MM & Acc$_{n}$\\ \midrule
\multirow{2}{*}{\STAB{\rotatebox[origin=c]{90}{SoTA}}} 
& MTMSN              & $86.5$   & $89.4$     & $3$ &  81.3     \\
& NeRd               & $73$     & $76.6$     & $2$ &    62.3    \\\cmidrule(l){2-6} 
\multirow{2}{*}{\STAB{\rotatebox[origin=c]{90}{Ours}}} 
& Decomp$_{G}$    & $78.8$   & $85.1$    & 1      & -       \\
& Decomp$_{L}$ & $74.4 \pm 2.4$     & $79.9 \pm 2.6$    & 1 & $64$\\ \bottomrule
\end{tabular}}
\caption{Accuracy of models for subtraction questions.  We report accuracy on \emph{clean} evaluation set (52 questions) in Acc$_{c}$, accuracy after omitting 5 mislabeled questions in the second column (Acc$_{c}^{-}$) and specify how many of these Mislabeled questions Match the prediction in the \#MM.  The last column ( Acc$_{n}$) reports accuracy on the \emph{noisy} evaluation set (892 questions). Learned Decompositions (Decomp$_{L}$) are averaged over 3 random seeds in pointer model training. }
\label{table:subtractionresults}
\end{table}

Table \ref{table:subtractionresults} shows the accuracy of each of the models on the subtraction evaluation sets. 
Since the result is a number, accuracy is evaluated as an exact match between the predicted answer and its label.
We compare our approach with the state-of-the-art; MTMSN \cite{hu-etal-2019-multi}, the best performing model with specialized modules; and NeRd \cite{Chen2020Neural}, the most recent work based on program induction. These were evaluated on the original questions in subtraction evaluation set.
For our work, we evaluate two different variations: We run the pipeline on the gold decompositions that have been manually rewritten, and automatically-decomposed questions generated by our approach, using BERT single-hop RC described in section \ref{sec:exp}. For both gold-decompositions and learned-decompositions we get promising results that are on par with the state-of-the-art on this dataset.

When investigating the mistakes that our approach makes on the \emph{clean} set, we find that many of the mistakes are actually due to incorrect labels. The gold answer (or label) does not match the correct answer for a certain question. To validate this, we check the entire \emph{clean} evaluation set and manually label each question. We find that 5 of the 52 questions are incorrectly annotated, one of these questions is actually invalid as the information needed to answer it does not exist in the passage. 
To better understand the effect of this, we discard incorrectly labeled examples and report accuracy in the second column of Table~\ref{table:subtractionresults}. We also report the number of predictions matching the incorrect label. 
The primary set of \textit{true} mistakes our model makes are due to some questions needing further decomposition, eliciting common-sense knowledge, or because they are not \textit{subtraction} questions, i.e. can be classified as MTMSN's \texttt{Negation} rather than \texttt{Diff} module.

NeRd fails on 3 questions that MTMSN and our approach got correctly because it could not produce a valid program to be evaluated. It also failed on 2 of the \texttt{Negation} question that our approach failed on, not because it was not able to address those kinds of question, but because the attention mechanism ignored a condition in the question ``18 or over". Surprisingly, NeRd failed on both questions that necessitate nested processing, even though the architecture allows for compositionality. The remaining failure cases are due to choosing incorrect operands for the difference, but it is not clear why the model made those choices.

\paragraph{Discussion}
We find that our approach is promising; it is interpretable and requires little training data when compared to previous approaches, without compromising performance. Steps to arrive at an answer are explicit, and we can interpret each of the retrieved operands by their associated subquestions. Figure~\ref{fig:model-overviewl} shows an example of this for subtraction questions. MTMSN indicates which module was used, but it does not show what led to this particular choice of the arithmetic expression. Likewise, NeRD shows the program necessary to find the answer, but there is no indication on why the operands of each function were chosen.

The only training data needed was a small subset (200 examples) to train the pointer model, and in the future we need some data to train reasoning type classifier and other templates' pointer models. This  comes in contrast to the exhaustive search needed to find all possible derivations to reach an answer for all questions in the training set (77.4k examples).
Reasoning Templates retrieve operands for the subtraction operation by answering subquestions that refer to a particular number, making it more robust to noise in the annotation.
We started by focusing on the subtraction template, because it is the most prevalent numerical reasoning type (with an estimated proportion of 29\% of all questions \cite{Dua2019DROP}). However, this approach can be similarly extended to other reasoning types by defining a template for each, such as \textit{date-difference} or \textit{addition}.

We believe that such reasoning templates would be able to answer compositional questions with its recursive \emph{decomposition} component. While this exploration is left for future research, we believe it is useful it outline how we expect it to handle compositionality.
Recall from Figure~\ref{fig:model-overviewl} that input questions are passed to a classifier that selects which template to apply, one of the classes decides if the question is single-span and should be passed on to single-hop RC directly. Decomposed questions should also be passed through this classifier to determine if they need further decomposition. 
\begin{figure}[ht]
    \centering
\includegraphics[width=\columnwidth]{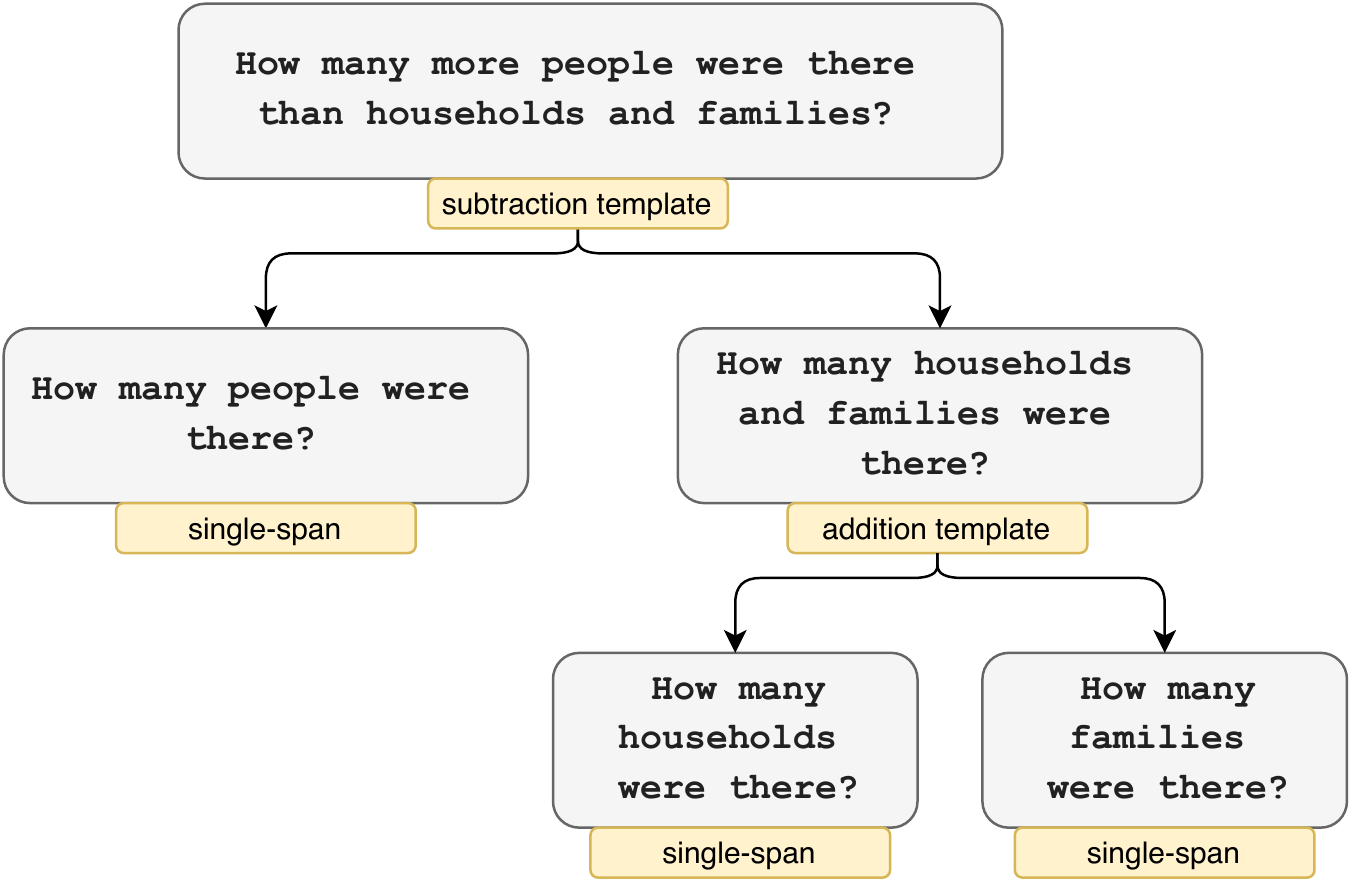}
    \caption{An example of how questions are further decomposed to facilitate compositionality.}
    \label{fig:compositionality}
\end{figure}

Figure~\ref{fig:compositionality} shows an example where the second subquestion involves another reasoning task. After further decomposing it to single-span extraction questions and finding the solution to the addition operation, that answer would be passed to the previous task. This recursive processing should ideally allow for compositionality. 

After building the entire pipeline we expect mistakes like nested operations and mis-classified \texttt{Negation} types to be rectified, boosting performance further.
One challenge we wish to overcome is the engineering bottleneck involved in crafting each of the templates. Future work would explore methods that learn to construct these the templates.

\section{Conclusion}
\label{sec:conc}
We propose using Reasoning Templates for tackling reading comprehension tasks that involve reasoning over multiple paragraphs. 
We show that this approach is competitive with state of the art models on a subset of DROP's subtraction questions, while requiring much less training data and providing better visibility of the model's decision making. In future work, we plan on extending to further templates and investigate how to learn templates instead of working from a predefined set.

\section*{Acknowledgements}
This research has been supported by a scholarship from King Saud University. We thank our anonymous mentor and reviewers for their constructive comments and suggestions.  
\clearpage
\bibliography{anthology,eacl2021}

\begin{thebibliography}{18}
\expandafter\ifx\csname natexlab\endcsname\relax\def\natexlab#1{#1}\fi

\bibitem[{Andor et~al.(2019)Andor, He, Lee, and
  Pitler}]{andor-etal-2019-giving}
Daniel Andor, Luheng He, Kenton Lee, and Emily Pitler. 2019.
\newblock \href {https://doi.org/10.18653/v1/D19-1609} {Giving {BERT} a
  calculator: Finding operations and arguments with reading comprehension}.
\newblock In \emph{Proceedings of the 2019 Conference on Empirical Methods in
  Natural Language Processing and the 9th International Joint Conference on
  Natural Language Processing (EMNLP-IJCNLP)}, pages 5947--5952, Hong Kong,
  China. Association for Computational Linguistics.

\bibitem[{Chen et~al.(2020)Chen, Liang, Yu, Zhou, Song, and
  Le}]{Chen2020Neural}
Xinyun Chen, Chen Liang, Adams~Wei Yu, Denny Zhou, Dawn Song, and Quoc~V. Le.
  2020.
\newblock \href {https://openreview.net/forum?id=ryxjnREFwH} {Neural symbolic
  reader: Scalable integration of distributed and symbolic representations for
  reading comprehension}.
\newblock In \emph{International Conference on Learning Representations}.

\bibitem[{Devlin et~al.(2019)Devlin, Chang, Lee, and
  Toutanova}]{devlin2018bert}
Jacob Devlin, Ming-Wei Chang, Kenton Lee, and Kristina Toutanova. 2019.
\newblock \href {https://doi.org/10.18653/v1/N19-1423} {{BERT}: Pre-training of
  deep bidirectional transformers for language understanding}.
\newblock In \emph{Proceedings of the 2019 Conference of the North {A}merican
  Chapter of the Association for Computational Linguistics: Human Language
  Technologies, Volume 1 (Long and Short Papers)}, pages 4171--4186,
  Minneapolis, Minnesota. Association for Computational Linguistics.

\bibitem[{Dua et~al.(2019)Dua, Wang, Dasigi, Stanovsky, Singh, and
  Gardner}]{Dua2019DROP}
Dheeru Dua, Yizhong Wang, Pradeep Dasigi, Gabriel Stanovsky, Sameer Singh, and
  Matt Gardner. 2019.
\newblock {DROP}: A reading comprehension benchmark requiring discrete
  reasoning over paragraphs.
\newblock In \emph{Proc. of NAACL}.

\bibitem[{Gupta et~al.(2020)Gupta, Lin, Roth, Singh, and
  Gardner}]{Gupta2020Neural}
Nitish Gupta, Kevin Lin, Dan Roth, Sameer Singh, and Matt Gardner. 2020.
\newblock \href {https://openreview.net/forum?id=SygWvAVFPr} {Neural module
  networks for reasoning over text}.
\newblock In \emph{International Conference on Learning Representations}.

\bibitem[{Honnibal et~al.(2020)Honnibal, Montani, Van~Landeghem, and
  Boyd}]{spacy2}
Matthew Honnibal, Ines Montani, Sofie Van~Landeghem, and Adriane Boyd. 2020.
\newblock \href {https://doi.org/10.5281/zenodo.1212303} {{spaCy:
  Industrial-strength Natural Language Processing in Python}}.

\bibitem[{Hu et~al.(2019)Hu, Peng, Huang, and Li}]{hu-etal-2019-multi}
Minghao Hu, Yuxing Peng, Zhen Huang, and Dongsheng Li. 2019.
\newblock \href {https://doi.org/10.18653/v1/D19-1170} {A multi-type multi-span
  network for reading comprehension that requires discrete reasoning}.
\newblock In \emph{Proceedings of the 2019 Conference on Empirical Methods in
  Natural Language Processing and the 9th International Joint Conference on
  Natural Language Processing (EMNLP-IJCNLP)}, pages 1596--1606, Hong Kong,
  China. Association for Computational Linguistics.

\bibitem[{Huang et~al.(2019)Huang, Le~Bras, Bhagavatula, and
  Choi}]{huang-etal-2019-cosmos}
Lifu Huang, Ronan Le~Bras, Chandra Bhagavatula, and Yejin Choi. 2019.
\newblock \href {https://doi.org/10.18653/v1/D19-1243} {Cosmos {QA}: Machine
  reading comprehension with contextual commonsense reasoning}.
\newblock In \emph{Proceedings of the 2019 Conference on Empirical Methods in
  Natural Language Processing and the 9th International Joint Conference on
  Natural Language Processing (EMNLP-IJCNLP)}, pages 2391--2401, Hong Kong,
  China. Association for Computational Linguistics.

\bibitem[{Khashabi et~al.(2018)Khashabi, Chaturvedi, Roth, Upadhyay, and
  Roth}]{khashabi-etal-2018-looking}
Daniel Khashabi, Snigdha Chaturvedi, Michael Roth, Shyam Upadhyay, and Dan
  Roth. 2018.
\newblock \href {https://doi.org/10.18653/v1/N18-1023} {Looking beyond the
  surface: A challenge set for reading comprehension over multiple sentences}.
\newblock In \emph{Proceedings of the 2018 Conference of the North {A}merican
  Chapter of the Association for Computational Linguistics: Human Language
  Technologies, Volume 1 (Long Papers)}, pages 252--262, New Orleans,
  Louisiana. Association for Computational Linguistics.

\bibitem[{Kusner et~al.(2015)Kusner, Sun, Kolkin, and
  Weinberger}]{Kusner-2015-WMD}
Matt~J. Kusner, Yu~Sun, Nicholas~I. Kolkin, and Kilian~Q. Weinberger. 2015.
\newblock From word embeddings to document distances.
\newblock In \emph{Proceedings of the 32nd International Conference on
  International Conference on Machine Learning - Volume 37}, ICML'15, page
  957–966. JMLR.org.

\bibitem[{Min et~al.(2019)Min, Zhong, Zettlemoyer, and
  Hajishirzi}]{min-etal-2019-multi}
Sewon Min, Victor Zhong, Luke Zettlemoyer, and Hannaneh Hajishirzi. 2019.
\newblock \href {https://doi.org/10.18653/v1/P19-1613} {Multi-hop reading
  comprehension through question decomposition and rescoring}.
\newblock In \emph{Proceedings of the 57th Annual Meeting of the Association
  for Computational Linguistics}, pages 6097--6109, Florence, Italy.
  Association for Computational Linguistics.

\bibitem[{Rajpurkar et~al.(2018)Rajpurkar, Jia, and Liang}]{rajpurkar2018know}
Pranav Rajpurkar, Robin Jia, and Percy Liang. 2018.
\newblock \href {https://doi.org/10.18653/v1/P18-2124} {Know what you don{'}t
  know: Unanswerable questions for {SQ}u{AD}}.
\newblock In \emph{Proceedings of the 56th Annual Meeting of the Association
  for Computational Linguistics (Volume 2: Short Papers)}, pages 784--789,
  Melbourne, Australia. Association for Computational Linguistics.

\bibitem[{Rajpurkar et~al.(2016)Rajpurkar, Zhang, Lopyrev, and
  Liang}]{rajpurkar-etal-2016-squad}
Pranav Rajpurkar, Jian Zhang, Konstantin Lopyrev, and Percy Liang. 2016.
\newblock \href {https://doi.org/10.18653/v1/D16-1264} {{SQ}u{AD}: 100,000+
  questions for machine comprehension of text}.
\newblock In \emph{Proceedings of the 2016 Conference on Empirical Methods in
  Natural Language Processing}, pages 2383--2392, Austin, Texas. Association
  for Computational Linguistics.

\bibitem[{Ran et~al.(2019)Ran, Lin, Li, Zhou, and Liu}]{ran-etal-2019-numnet}
Qiu Ran, Yankai Lin, Peng Li, Jie Zhou, and Zhiyuan Liu. 2019.
\newblock \href {https://doi.org/10.18653/v1/D19-1251} {{N}um{N}et: Machine
  reading comprehension with numerical reasoning}.
\newblock In \emph{Proceedings of the 2019 Conference on Empirical Methods in
  Natural Language Processing and the 9th International Joint Conference on
  Natural Language Processing (EMNLP-IJCNLP)}, pages 2474--2484, Hong Kong,
  China. Association for Computational Linguistics.

\bibitem[{Reddy et~al.(2019)Reddy, Chen, and Manning}]{reddy-etal-2019-coqa}
Siva Reddy, Danqi Chen, and Christopher~D. Manning. 2019.
\newblock \href {https://doi.org/10.1162/tacl_a_00266} {{C}o{QA}: A
  conversational question answering challenge}.
\newblock \emph{Transactions of the Association for Computational Linguistics},
  7:249--266.

\bibitem[{Wolf et~al.(2020)Wolf, Debut, Sanh, Chaumond, Delangue, Moi, Cistac,
  Rault, Louf, Funtowicz, Davison, Shleifer, von Platen, Ma, Jernite, Plu, Xu,
  Scao, Gugger, Drame, Lhoest, and Rush}]{Wolf2019HuggingFacesTS}
Thomas Wolf, Lysandre Debut, Victor Sanh, Julien Chaumond, Clement Delangue,
  Anthony Moi, Pierric Cistac, Tim Rault, Rémi Louf, Morgan Funtowicz, Joe
  Davison, Sam Shleifer, Patrick von Platen, Clara Ma, Yacine Jernite, Julien
  Plu, Canwen Xu, Teven~Le Scao, Sylvain Gugger, Mariama Drame, Quentin Lhoest,
  and Alexander~M. Rush. 2020.
\newblock \href {https://www.aclweb.org/anthology/2020.emnlp-demos.6}
  {Transformers: State-of-the-art natural language processing}.
\newblock In \emph{Proceedings of the 2020 Conference on Empirical Methods in
  Natural Language Processing: System Demonstrations}, pages 38--45, Online.
  Association for Computational Linguistics.

\bibitem[{Wolfson et~al.(2020)Wolfson, Geva, Gupta, Gardner, Goldberg, Deutch,
  and Berant}]{Wolfson2020Break}
Tomer Wolfson, Mor Geva, Ankit Gupta, Matt Gardner, Yoav Goldberg, Daniel
  Deutch, and Jonathan Berant. 2020.
\newblock Break it down: A question understanding benchmark.
\newblock \emph{Transactions of the Association for Computational Linguistics}.

\bibitem[{Yang et~al.(2018)Yang, Qi, Zhang, Bengio, Cohen, Salakhutdinov, and
  Manning}]{yang2018hotpotqa}
Zhilin Yang, Peng Qi, Saizheng Zhang, Yoshua Bengio, William~W. Cohen, Ruslan
  Salakhutdinov, and Christopher~D. Manning. 2018.
\newblock {HotpotQA}: A dataset for diverse, explainable multi-hop question
  answering.
\newblock In \emph{Conference on Empirical Methods in Natural Language
  Processing ({EMNLP})}.

\end{thebibliography}
\bibliographystyle{acl_natbib}

\appendix
\section{Experimental Settings}
\subsection{Model Settings}
We use the final layer of BERT$_{\text{LARGE}}$ \cite{devlin2018bert} to produce contextualized embeddings used for span extraction 
fine-tuned to extract 4 pointers.\footnote{We build upon the implementation of \newcite{min-etal-2019-multi}} We use Adam optimizer with learning rate of $5e-5$ and warm-up over the first $10\%$ steps to train. Loss function is calculated with cross-entropy. Training batch size is 20 examples. We train three models with different random seeds and report average performance over these.

\subsection{SoTA Comparison}
We report the accuracy of MTMSN \cite{hu-etal-2019-multi} and NeRd \cite{Chen2020Neural} on the two subtraction evaluation sets in Table~\ref{table:subtractionresults}. 
For MTMSN, we use the pre-trained \texttt{MTMSN\_LARGE} model published on their \href{https://github.com/huminghao16/MTMSN}{github} page. Code and model checkpoints for NeRd were shared in email communication with the authors in June, 2020.

\section{Evaluating on a larger dataset}
\label{appx-larger}
We started evaluating this work on the smaller, \emph{clean}, dataset of 52 questions that has been manually curated. To validate that this sample is representative of subtraction questions in the DROP devset, we worked to heuristically identify relevant questions. We started with the same 2 steps involved in the manual curation, filter the devset ($9536$ questions) for questions that have `number' as answer type (leaves $5850$ questions)  and contain comparative adjectives or adverbs. This leaves us with a subset of $1386$ questions. The above conditions cover more questions than we are interested in, e.g. `How many people are 18 or older?'. We refine the second condition to exclude sentences where the \texttt{JJR|RBR} tokens are preceded with an \emph{or}, this omits $146$ more samples. We proceed by passing these through our pipeline. Below is a summary of failure cases of the different components of our approach:
\begin{enumerate}[a.]
    \item $1$ sample did not produce valid pointers (used [SEP] token which is BERT-specific).
    \item $22$ samples did not produce valid decomposition. This is due to issues in mismatching tokenization between the pointer model and the subquestion generation function. The function used to map pointers between the two tokenizers did not generalize to the cases here. Examples of these are [`80', `-', `yard'] and [`80-yard'].
    \item $28$ samples did not pass through BERTQA successfully, as they exceeded the sequence length ($512$). 
\end{enumerate}
Of the remaining $1189$ questions that were processed successfully, we get $55.9\%$ correctly. This still includes questions which are not covered by our subtraction template. We proceed in two ways: First, we filter out questions that MTMSN predicted not to be \texttt{addition\_subtraction}. This leaves $1106$ questions with $59.3\%$ accuracy. The alternative is to filter questions based on their start trigrams, which gives a more relevant set of questions. Of the $1189$ questions, $892$ start with the phrases `How many more', `How many fewer', and `How many less'. Our model answers $64\%$ of these correctly.

\end{document}